\documentclass{article}
\usepackage{ijcai18}

\usepackage{times}
\usepackage{xcolor}
\usepackage{soul}
\usepackage{graphicx}
\usepackage[utf8]{inputenc}
\usepackage[small]{caption}
\let\ACMmaketitle=\maketitle
\renewcommand{\maketitle}{\begingroup\let\footnote=\thanks \ACMmaketitle\endgroup}

\title{Automatic Target Recovery for Hindi-English Code Mixed Puns\footnote{This work was presented at 1st Workshop on Humanizing AI (HAI) at IJCAI'18 in Stockholm, Sweden.}}

\author{
Srishti Aggarwal$^1$, 
Kritik Mathur$^2$, 
Radhika Mamidi$^1$
\\ 
$^1$ LTRC, KCIS, IIIT Hyderabad \\
$^2$ Manipal Institute of Technology\\
srishti.aggarwal@research.iiit.ac.in,
krt.mat@gmail.com,
radhika.mamidi@iiit.ac.in
}

\begin{document}

\maketitle

\begin{abstract}

In order for our computer systems to be more human-like, with a higher emotional quotient, they need to be able to process and understand intrinsic human language phenomena like humour. In this paper, we consider a subtype of humour - puns, which are a common type of wordplay-based jokes. In particular, we consider code-mixed puns which have become increasingly mainstream on social media, in informal conversations and advertisements and aim to build a system which can automatically identify the pun location and recover the target of such puns. We first study and classify code-mixed puns into two categories namely intra-sentential and intra-word, and then propose a four-step algorithm to recover the pun targets for puns belonging to the intra-sentential category. Our algorithm uses language models, and phonetic similarity-based features to get the desired results. We test our approach on a small set of code-mixed punning advertisements, and observe that our system is successfully able to recover the targets for 67\% of the puns.
\end{abstract}

\section{Introduction}

Humour is one of the most complex and intriguing phenomenon of the human language. It exists in various forms, across space and time, in literature and culture, and is a valued part of human interactions. 
Puns are one of the simplest and most common forms of humour in the English language. They are also one of the most widespread forms of spontaneous humour \cite{ritchie2004linguistic} and have found their place in casual conversations, literature, online comments, tweets and advertisements \cite{2018arXiv180405398M,TANAKA199291}.
Puns are a hugely versatile and commonly used literary device and it is essential to include them in any comprehensive approach to computational humour.

In this paper, we consider Hindi-English code-mixed puns and aim to automatically recover their targets. The target of a pun is its phonologically similar counterpart, the relationship to which and whose resolution (recovery) in the mind of the listener/hearer induces humour. For example, in the pun ``The life of a patient of hypertension is always at steak." the  word ``steak" is the pun with target ``stake".

With India being a diverse linguistic region, there is an ever increasing usage of code-mixed Hindi-English language (along with various others) because bilingualism and even multilingualism are quite common. Consequently, we have also seen an increase in the usage of code-mixed language in online forums, advertisements etc. Code-mixed humour, especially puns have become increasingly popular because being able to use the same punning techniques but with two languages in play has opened up numerous avenues for new and interesting wordplays. With the increasing popularity and acceptance for the usage of code-mixed language, it has become important that computers are also able to process it and even decipher complex phenomena like humour. 

Traditional Word Sense Disambiguation (WSD) based methods cannot be used in target recovery of code-mixed puns, because they are no longer about multiple senses of a single word but about two words from two different languages. Code-switching comes with no markers, and the punning word may not even be a word in either of the languages being used. Sometimes words from the two languages can be combined to form a word which only a bilingual speaker would understand. Hence, this task on such data calls for a different set of strategies altogether. We approach this problem in two parts. First, we analyze the types of structures in code-mixed puns and classify them into two categories namely intra-sequential and intra-word. Second, we develop a four stage pipeline to achieve our goal - Language Identification, Pun Candidate Identification, Context Lookup and Phonetic Distance Minimization. We then test our approach on a small dataset and note that our method is successfully able to recover targets for a majority of the puns.

To the best of our knowledge, this is a first attempt at dealing with code-mixed puns. The outline of the paper is as follows: Section 2 gives a brief description of the background and prior work on puns - both in the field of linguistics and in the field of computational humour, along with a brief introduction to the field of code-mixing. Section 3 defines our problem statement, our classification model on code-mixed puns, the dataset we use to test our approach, and our proposed model for the task of automatic target recovery of Hindi-English code-mixed puns. In Section 4, we analyse the performance of our model on a set of puns, and discuss the various error cases. Finally, we conclude in Section 5 with a review of our research contributions and an outline of our plans for future work.
 
\section{Background and Related Work}
\subsection{Puns}
\subsubsection{Linguistic Studies}
Puns are a form of wordplay jokes in which one sign (e.g. a word or a phrase) suggests two or more meanings by exploiting polysemy, homonymy, or phonological similarity to another sign, for an intended humorous or rhetorical effect \cite{doi:10.4324/9781315731162.ch7}. Puns where the two meanings share the same pronunciation are known as \emph{homophonic} or \emph{perfect}\footnote{Example of a perfect pun: ``The tallest building in town is the library — it has thousands of stories!''} puns, while those relying on similar but non-identical sounding words are known as \emph{heterophonic} \cite{hempelmann2003paronomasic} or \emph{imperfect}\footnote{An example of an imperfect puns: ``with fronds like these, who needs anemones'' \cite{zwicky1986imperfect}} puns \cite{zwicky1986imperfect}. In this paper, we study automatic target recoverability of English-Hindi code mixed puns - which are more commonly imperfect puns, but may also be perfect puns in some cases.

Zwicky and Zwicky \shortcite{zwicky1986imperfect}, Sobkowiak \shortcite{sobkowiak1991metaphonology} extensively studied various phonological variations in imperfect puns such as strong asymmetry in phoneme substitution. They note that puns show more frequent changes in vowels than in consonants because of their smaller role in target recoverability.

\subsubsection{Puns in Computational humour}

Puns have received attention  in the field of computational humour, both in generation of puns and their understanding.

\textbf{Generation:} One of the earliest attempts at generating humour was by Lessard and Levin \shortcite{lessard1992computational}, when they built an antonym-based system to generate \emph{Tom Swifties}\footnote{ https://en.wikipedia.org/wiki/Tom\_Swifty }. Since then, we have seen various other attempts at the task with different strategies. JAPE was a system which exploited framing and phonetic relationships to automatically generate funny punning riddles, or more specifically phonologically ambiguous riddles, having noun phrase punchlines \cite{binsted1994implemented}. Venour \shortcite{venour1999computational} built a system which generated HCPPs (Homonym Common Phrase Pun), simple 2 sentence puns based on associations between words occurring in common phrases. WisCraic was a system built by McKay \shortcite{mckay2002generation}, which generated simple one-sentence puns based on semantic associations of words. Valitutti et al. \shortcite{valitutti2008textual} attempted to automatically generate advertisements by punning on familiar expressions, with an affective connotation.

\textbf{Identification and understanding:} Hempelmann \shortcite{hempelmann2003paronomasic} studied target recoverability, arguing that a good model for it provides necessary groundwork for effective automatic pun generation. He worked on a theory which models prominent factors in punning such as phonological similarity and studied how these measures could be used to evaluate possible imperfect puns given an input word and a set of target words. 

Yokogawa \shortcite{yokogawa2002japanese} analyzed ungrammatical Japanese puns and generated target candidates by replacing ungrammatical parts of the sentence by similar expressions. Taylor and Mazlack \shortcite{taylor2004computationally} worked on computational recognition of word-play in the restricted domain of Knock-Knock jokes. Jaech et al. \shortcite{jaech2016phonological} developed a computational model for target recovery of puns using techniques for automatic speech recognition, and learned phone edit probabilities in puns. Miller and Gurevych \shortcite{Miller2015AutomaticDO}, Miller et al.\shortcite{miller2017semeval} describe different methods on pun identification and disambiguation. Word Sense Disambiguation (WSD) based techniques are most common among the methods used.

To the best of our knowledge no prior work has been attempted on code-mixed puns. 

\subsection{Code-mixing}
Code-mixing is the mixing of two or more languages or language varieties. Code-mixing is now recognized as a natural part of bilingual and multilingual language use. Significant linguistic efforts have been made to understand the sociological and conversational necessity behind code-switching \cite{auer1984bilingual}; for example, to understand whether it is an act of identity in a social group, or a consequence of a lack of competence in either of the languages. These papers distinguish between inter-sentence, intra-sentence and intra-word code mixing.

Different types of language mixing phenomena have been discussed and defined by several linguists, with some making clear distinctions between phenomena based on certain criteria, while others use `code-mixing’ or `code-switching’ as umbrella terms to include any type of language mixing — see, e.g., Muysken \shortcite{muysken1995code} or Gafaranga and Torras \shortcite{gafaranga2002interactional}. In this paper, we use both these terms ‘code-mixing’ and `code-switching' interchangeably.

Coming to the work on automatic analysis of code-mixed languages, there have been studies on detecting code mixing in spoken language as well as different types of short texts, such as information retrieval queries \cite{gottron2010comparison}, SMS messages \cite{farrugia2004tts,rosner2007tagging}, social media data \cite{Barman14code-mixing:a} and online conversations \cite{nguyen2013word}. These scholars have carried out experiments for the task of language identification using language models, dictionaries, logistic regression classification, Conditional Random Fields, SVMs, and noted that approaches using contextual knowledge were most robust. King and Abney \shortcite{king2013labeling} used weakly semi-supervised methods to perform word-level language identification. 

We however, use a dictionary based approach for the language identification task. While working with puns, ambiguity in language identification can be an important marker for identifying the pun, so it is more important for us to recognize all possible ambiguities rather than picking just one depending on probabilities. This ability to recognize ambiguities, and the simplicity of a dictionary-based language identification model makes it suited for this task.

\section{Methodology}
We focus on the task of automatically disambiguating or recovering Hindi-English code mixed puns. For this purpose, it is first necessary to understand what these puns are.

\subsection{Classification}
For the purposes of this research, we only consider puns where the ambiguity or the wordplay lies in the code-switching i.e, the pun word and its target are from different languages. For example the pun "Rivers can't hear because \emph{woh behri hoti hai}." is a sentence with the pun being \emph{behri} (meaning deaf) and its target being \emph{beh rahi} (meaning flowing). Here, while the sentence is code-mixed, the pun word and the target both belong to the same language. We do not consider such puns for the present study. 


We analyze the structure of code-mixed puns with the pun word and its target belonging to different languages and propose two broad categories to classify them in - puns where the code-mixing is intra-sentential and the other where it is intra-word. Both these categories are explained below, while we evaluate only on the former category.

\subsubsection{Intra-Sentential Code-Mixed Puns}
Intra-sentential code-mixing is where code-switching occurs within a sentence. Here, the language varies at the word level. Also, each word of the sentence belongs to one or the other language. Table \ref{Example-1} gives examples of puns belonging to this category.
\\
\begin{table}[h!]
\centering
\begin{tabular}{lll}
\hline
\\
\textbf{Pun\textsubscript{1}}&:& Grand \emph{Salaam}\\
Translation&:& Grand Salute\\
Pun Location&:& \emph{Salaam}\\
Target&:& Grand Slam\\
\\
\hline
\\
\textbf{Pun\textsubscript{2}}&:&\emph{Phir bhi} zeal \emph{hai Hindustani}\\
Translation&:& The zeal is still Indian\\
Pun Location&:&zeal\\
Target&:& \emph{Phir bhi dil hai Hindustani}\\
Translation (Target)&:& The heart is still Indian.\\
\\
\hline
\end{tabular}
\caption{Examples of intra-sentential code-mixed puns}
\label{Example-1}
\end{table}

\subsubsection{Intra-Word Code Mixed Puns}
In this category, code mixing is present within a word. New words are formed using Portmanteau\footnote{https://en.wikipedia.org/wiki/Portmanteau} or Blending where two or more syllables/phonemes from different languages are blended together to form a single word, resulting in a word which is phonetically similar to the target word. Table \ref{Example-2} illustrates examples of intra-word code-mixed puns.

\begin{table}[h!]
\centering
\begin{tabular}{lll}
\hline
\\
\textbf{Pun\textsubscript{3}}&: &\emph{Facebhukh} with Amul, Mark\\
Translation&: &Face hunger with Amul, Mark\\
Pun Location&: &\emph{Facehbhukh}\\
Target&:& Facebook with Amul, Mark\\
\\
\hline
\\
\textbf{Pun\textsubscript{4}}&: &\emph{Rajnitea?}\\
Translation&:& Rajni, tea?\\
Pun Location&: &\emph{Rajnitea}\\
Target&: &\emph{Rajneeti}\\
Translation(target)&: & Politics\\
\\
\hline
\end{tabular}
\caption{Examples of intra-word code-mixed puns}
\label{Example-2}
\end{table}

\subsection{Dataset}
Most puns we hear or use in everyday conversations are rarely recorded. One of the most common resources to find recorded puns are advertisements, for example the highly creative and frequently released Amul advertisements in India \cite{2018arXiv180405398M}. Most of these are \emph{contextually integrated} \cite{ritchie2004linguistic} with an image. While such puns may lose their humour out of context, it is still possible to recover their targets, so using these does not affect our task in any way

To create a dataset to test our model on, we collected 518 advertisements released by Amul in the years 2014, 2015, 2017 and 2018, from their official web page\footnote{http://www.amul.com/m/amul-hits}. Of these, 333 were puns, including 121 code-mixed puns as defined in Section 3.1. We  extracted the text of these 121 code-mixed puns and asked 3 people to disambiguate them, given just the advertisement text. All three annotators were university students in 22-23 years age group, native Hindi speakers with bilingual fluency in English. The annotators were asked to identify the location of the pun in each of the advertisements and write down the target of the pun. Any disagreements between annotators were resolved by mutual discussion. 

In a few cases where puns were identified to have multiple targets, we kept all such possibilities in our dataset. A few puns were identified to be non-recoverable because of the lack of contextual knowledge, while a few puns had multiple pun locations. We removed both these types from our dataset, which left us with 110 puns.

Finally, we divided these 110 annotated puns into the two categories as defined in Section 3.1 thereby getting 51 advertisements categorized as intra-sentential code-mixed puns, and the rest as intra-word code-mixed puns. We use the former as our test data.

\begin{figure*}[t!]
\vspace{5mm}
\centering
\includegraphics[width = 16cm]{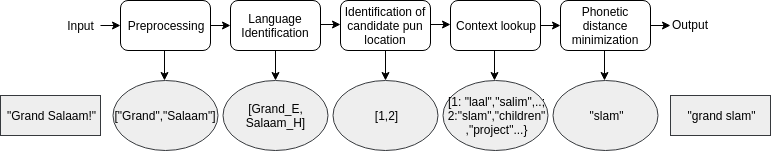}
\caption{This figure illustrates, taking Pun\textsubscript{1} as example, our model and the 4 major steps it comprises: 1. Language Identification, 2. Identification of Candidate Pun Locations, 3. Context Lookup and 4. Phonetic Distance minimization. }
\end{figure*}

\subsection{Model}

For preprocessing the text we give as input to our system, we first tokenize the advertisement text using NLTK's \cite{bird2009natural} tokenizer and remove all punctuations. We then give the resultant tokens as input to our model, which is a 4 step process as described below:

\subsubsection{Step 1: Language Identification}
At this step, we aim to identify the language of each of the tokens in the input text by classifying them into one of the 5 categories: English, Hindi, Named Entity (NE), Out of Vocabulary (OOV), or Ambiguous (words that could belong to both English and Hindi).

We use a dictionary-based lookup method to classify a word in English or Hindi. Since the input is in Roman script, to recognize Hindi words, we use a list of 30k transliterated Hindi words in Roman to their Devanagari counterparts \cite{gupta2012mining}. For the English language, we collected news data from the archives of a leading Indian Newspaper, The Hindu\footnote{http://www.thehindu.com/archive/}. Data from 2012-2018 under the tags National, International, Sports, Cinema, Television was collected, amounting to 12,600 articles with 200k sentences and around 38k unique words. We use this data to build an English dictionary. Also, we used NLTK's \cite{bird2009natural} Named Entity Recognition module on the same data to get a dictionary of Named Entities. 

We first try to classify all tokens as English, Hindi and NE using these dictionaries. Then, words which are found in both English and Hindi are marked as Ambiguous. The words which do not fall into any of these are classified as OOV.

\subsubsection{Step 2: Identification of Candidate Pun Locations} 
We now identify all possible punning locations in the text. For this, we consider words on the boundaries of language change as candidates for pun locations. Then, all NEs and OOV words are added to the list of pun candidates as well. Third, if any Ambiguous words exist in the text, we consider it once as English and once as Hindi for the next steps.

\subsubsection{Step 3: Context Lookup} 
In this step, we contextually lookup all the candidate locations using left context and right context to get a list of all words that may occur at that position. We use bi-gram language models we built using Knesser-Ney smoothing\footnote{we used the implementation from https://github.com/smilli/kneser-ney to build our language model} \cite{479394}. We used the data mentioned in the previous step to build the language model for English, and 100k sentences from Hindi monolingual data from \cite{kunchukuttan2017iit} to build the language models for English and Hindi respectively. As it is highly likely that the left and the right context at a pun location belong to different languages, we look at each of those separately instead of taking an intersection of the left and the right context.

\subsubsection{Step 4: Phonetic Distance Minimization}
Lastly, at each pun location, we calculate the similarity of the word at that location with all the words that can occur at that location depending on the context and pick the most similar words as the possible targets.

To compare words belonging to two different languages on a phonetic basis, we convert both of them to WX notation \cite{bharati1995natural}, which denotes a standard way to represent Indian languages in the Roman script. We transliterate our identified Hindi words from Devanagari to WX notation\footnote{We used the open source code available at https://github.com/ltrc/indic-wx-converter for this}. To convert English words to the same notation, we use the CMU phonetic dictionary \footnote{http://www.speech.cs.cmu.edu/cgi-bin/cmudict}, which uses a 39 phoneme set to represent North American pronunciations of English words. We build a mapping between this phoneme set and WX notation. Whenever there was no exact parallel between CMU pronouncing dictionary's notation and WX, we used the word's Indian English pronunciation to find the closest match.

Once we converted all to WX notation, we use a modified version of Levenshtein Distance \cite{Miller:2009:LDI:1822502} to find most similar words. In this normalized version of Levenshtein distance, we account for a few features like aspirations (for example, /p/,/ph/) which are non-phonemic in English, vowel elongations, rhyme, same beginning or ending sounds.

\begin{table}[h!]
\centering
\begin{tabular}{lll}
\hline
\\
\textbf{Pun\textsubscript{5}}&:&\emph{Aa} bail \emph{mujhe maar.}\\
Translation&:& Bail, come hit me.\\ 
Location&:&Bail\\
Target&:& \emph{Aa bail mujhe maar}\\
Translation (target)&:& Come bull, hit me.\\
\\
\hline
\\
\textbf{Pun\textsubscript{6}}&:& \emph{Doodh} Morning!\\
Translation&:&Milk Morning\\
Location&:&\emph{Doodh}, Morning\\
Target&:& Good Morning!\\
\\
\hline
\end{tabular}
\caption{Examples of puns successfully recovered by our system}
\label{Success-1}
\end{table}

In case of an OOV word, since it cannot be converted to WX notation due to non-availability of any phonetic transcription, we simply find the words with the least orthographic distance when written in Roman script, using a similar measure as used for phonetic distance with a few more normalizations (for example, considering 'w' and 'v' as similar).

\section{Results and discussion}

We test the model explained in the previous section on our test dataset described in Section 3.2 and note that this method is correctly able to recover targets for 34 out of these 51 puns, or around 67\% of the puns, which are very encouraging results for this complex task. Examples where the system performed successfully are given in Table \ref{Success-1}.

We do a thorough error analysis below for the cases our method fails for. 

\subsubsection{Error Analysis}
\begin{enumerate}
\item This method does not work when one word in the pun translates to multiple words in the target language. For example, pun\textsubscript{7} given in Table \ref{Error-1}, where a single word \emph{Vir}, which is the name of a person is a pun on ``We're". Our methodology is as of yet unable to recover such puns.

\begin{table}[h!]
\centering
\begin{tabular}{lll}
\hline
\\
\textbf{Pun\textsubscript{7}}&:&\emph{Vir} proud of you.\\
Translation&:& Vir (a name) proud of you\\ 
Location&:&\emph{Vir}\\
Target&:& We're proud of you.\\
\\
\hline
\end{tabular}
\caption{Example for error case 1, were a single pun word maps to more than one word in the target.}
\label{Error-1}
\end{table}

\item Our method fails when puns are based on pronunciation of abbreviations because we are not able to transliterate their pronunciation. For example Pun\textsubscript{8} given in Table \ref{Error-2}.

\begin{table}[h!]
\centering
\begin{tabular}{lll}
\hline
\\
\textbf{Pun\textsubscript{8}}&:& Greece, EU \emph{ro mat.}\\
Translation&:& Greece, EU don't cry.\\
Location&:& EU\\
Target&:& Greece, \emph{yun ro mat.}\\
Translation (target)&:& Greece, don't cry like this.\\
\\
\hline
\end{tabular}
\caption{Example for error case 2, where the pun is based on the pronunciation of an abbreviation.}
\label{Error-2}
\end{table}



\item Our model is also unable to recover the target, when the bigram doesn't really exist in the language model because it may be a new coinage or a very unusual phrase. In the example given in Table \ref{Error-3}, the target has ``slow food", but ``slow" is not an adjective that is normally associated with the noun ``food", and so the probability that it will occur in our language model is very low. Hence, our system fails to recover such targets.
\begin{table}[h!]
\centering
\begin{tabular}{lll}
\hline
\\
\textbf{Pun\textsubscript{9}}&:& Fast food ho ya \emph{sulu} food.\\
Translation&:& Whether it is fast food or 'sulu' food.\\
Location&:& \emph{'Sulu'}\\
Target&:& fast food ho ya slow food.\\
Translation (target)&:& Whether it is fast food or slow food.\\
\\
\hline
\end{tabular}
\caption{Example for error case 3, where the target does not exist in the language model.}
\label{Error-3}
\end{table}

\end{enumerate}
\section{Conclusion and Future work}
To conclude, in this paper, we present a first-ever work on target recovery code-mixed puns. We study various puns where the word-play is a result of code-switching, and classify them into 2 categories - puns with intra-sentential code mixing and those with intra-word code mixing. We then propose a methodology to recover the targets for puns belonging to the former category, using only monolingual language data. We test our proposed approach on a small manually annotated dataset, and we see that our system was able to successfully recover 67\% of the puns from the set. 

In the future, we want to perform a more comprehensive evaluation of this approach on a larger, more diverse set of puns. We want to improve and extend our approach to be able to recover intra-word code-mixed puns along with the intra-sentential ones that it handles right now. After that, the system should be extended to be able to recover all kinds of puns in code-mixed language, regardless of whether the pun itself is monolingual or code-mixed. 

\section*{Acknowledgements}
We thank the anonymous reviewers for their comments that helped improve this paper.

\bibliographystyle{named}
\bibliography{ijcai18}

\end{document}